\pdfoutput=1

\documentclass[11pt]{article}

\usepackage[]{acl}

\usepackage{times}
\usepackage{latexsym}
\usepackage{graphicx}

\usepackage[T1]{fontenc}

\usepackage[utf8]{inputenc}

\usepackage{microtype}

\definecolor{brilliantrose}{rgb}{1.0, 0.33, 0.64}
\newcommand{\rojo}[1]{{\color{black}{#1}}}

\newcommand{\jd}[1]{{\color{black}{#1}}}
\usepackage{todonotes}
\usepackage{booktabs}
\usepackage{threeparttable}

\title{Automatic Coding at Scale: Design and Deployment of a Nationwide System for Normalizing Referrals in the Chilean Public Healthcare System}

\author{Fabián Villena \\
  Center for Mathematical Modeling \\
  \& Department of Computer Sciences \\
  University of Chile \\
  \texttt{fabian.villena@uchile.cl} \\\And
  Matías Rojas \\
  Center for Mathematical Modeling \\
  University of Chile \\
  \texttt{matias.rojas.g@ug.uchile.cl} \\ \AND
  Felipe Arias \\
  Center for Mathematical Modeling \\
  University of Chile \\
  \texttt{felipe.arias.t@ug.uchile.cl} \\ \And
  Jorge Pacheco \\
  Dept. of Statistics and Health Information \\
  Chilean Ministry of Health \\
  \texttt{jorge.pacheco@minsal.cl} \\ \AND
  Paulina Vera \\
  Dept. of Statistics and Health Information \\
  Chilean Ministry of Health \\
  \texttt{paulina.vera@minsal.cl} \\ \And
  Jocelyn Dunstan \\
  Dept. Computer Science \& IMC \\
  Pontifical Catholic University of Chile \\
  \texttt{jdunstan@uc.cl} \\}

\begin{document}
\maketitle
\begin{abstract}
The disease coding task involves assigning a unique identifier from a controlled vocabulary to each disease mentioned in a clinical document. This task is relevant since it allows information extraction from unstructured data to perform, for example, epidemiological studies about the incidence and prevalence of diseases in a determined context. However, the manual coding process is subject to errors as it requires medical personnel to be competent in coding rules and terminology. In addition, this process consumes a lot of time and energy, which could be allocated to more clinically relevant tasks. These difficulties can be addressed by developing computational systems that automatically assign codes to diseases. In this way, we propose a two-step system for automatically coding diseases in referrals from the Chilean public healthcare system. Specifically, our model uses a state-of-the-art NER model for recognizing disease mentions and a search engine system based on Elasticsearch for assigning the most relevant codes associated with these disease mentions. The system's performance was evaluated on referrals manually coded by clinical experts. Our system obtained a MAP score of 0.63 for the subcategory level and 0.83 for the category level, close to the best-performing models in the literature. This system could be a support tool for health professionals, optimizing the coding and management process. Finally, to guarantee reproducibility, we publicly release the code of our models and experiments.
\end{abstract}

\section{Introduction}

The clinical text represents a significant proportion of patient's \rojo{health} records, commonly found in a non-structured format. These texts have particular challenges due to the extensive use of abbreviations, the variability of clinical language across medical specialties, and its restricted availability for privacy reasons \cite{dalianis2018}. Due to the complexity of its analysis, this data is commonly discarded in projects that seek to support clinical decision-making \cite{kong}.


Clinical coding involves mapping medical texts into codes using a controlled vocabulary consistent across different departments, hospitals, or even countries \cite{Dong2022}. The World Health Organization maintains an open, controlled vocabulary called the International Classification of Diseases (ICD), which is used in almost every country. Currently, the most widely used revision is the tenth (ICD-10) \cite{World_Health_Organization2015-tx}, and they are developing its eleventh revision, which will include not only diseases \cite{cie11}. 

\rojo{Regarding the Chilean public health system, the ICD-10 terminology is used for coding hospital discharges (morbidity coding by each healthcare provider) and deaths (mortality coding by the Ministry of Health).} Having patients' data normalized using these controlled vocabularies enables the ability to summarize information automatically and not deal with the noisiness of free-text data. The already-digested information from the normalized data empowers \rojo{data analysts who are not experts in NLP} to add more complex information into their workflows.



The Waiting Time Management System (SIGTE, in Spanish) contains electronic records of referrals from the Chilean Waiting List, which is the system that manages the high demand existent for consultation by specialists \cite{normaTecnica}. This data provided by 29 health services contain information about the medical diagnoses of patients but is not standardized \cite{baez2022procesamiento}.

As of November 2022, SIGTE recorded 25,374,491 waiting list referrals, of which 18,716,629 correspond to "new specialty referrals" and are associated with patient pathologies. Of these referrals, approximately 5,760,750 (30.7 \%) have an ICD-10 code. This calculation was performed by searching for a regular expression formatted as an ICD-10 code in the free-text diagnosis fields.

Clinical experts perform the disease coding task manually, which is not optimal for several reasons. Firstly, since this process is subject to errors, medical personnel must have significant competence in coding rules and a thorough knowledge of specialized terminologies, such as ICD\jd{, which also get updated frequently}. In other words, expert coding staff must be familiar with the clinical field, analytical and focused, and have fundamental skills for inspecting and analyzing highly specialized texts. In addition, manual coding is time-consuming \cite{YAN2022161}, which could be optimized by a support system, and this time could be used for other tasks relevant to clinical decision-making.

These difficulties can be efficiently addressed using computational systems capable of automatically performing the coding task using NLP. Currently, most automatic coding systems are based on an end-to-end architecture based on deep learning techniques. Although these systems have boosted the performance of several coding tasks, they cannot incorporate context-specific rules, such as code priority, medical assumptions, code definition, and synonyms.

In this work, we developed an automated disease coding system, thus being able to code the entire historical waiting list in Chile, identifying a total of 18,716,629 referrals. Our system is based on two steps; first, the automatic extraction of diseases is addressed using a state-of-the-art NER model, and then, using a search engine, the most probable code for each disease found is identified. Finally, we explored the potential applications derived from this system and studied in more depth the most frequent diseases in the country today.

\section{Related Work}

The disease coding task involves transforming clinical texts, commonly written by physicians in a non-structured format, into codes following medical terminologies. \jd{This is not an easy task since a medical ontology such as ICD in Spanish has
14,668 codes, an example of extreme multi-label classification \cite{barros2022divide}}. We have identified two major groups of computational methods proposed to solve this task; rule-based coding and neural network-based coding.

\subsection{Rule-based Models}

This approach involves designing hand-crafted rules to represent and simulate the flow that clinical experts follow when assigning codes. Most of the studies are based on using regular expressions and keywords to transform diseases found in the text into their respective codes. However, these methods are not feasible since manually capturing all the relations between texts and codes is time-consuming and complex.

Different approaches based on machine learning have been proposed to address this issue. In this way, features extracted from statistical models such as decision trees and support vector machines, among others, are incorporated into the manual rules \cite{coding-review, teng-et-al, Farkas2008AutomaticCO}. Another method is to create a list of synonyms of the original text to calculate a word distance with respect to the code descriptions of the terminology. Despite their disadvantages, these methods have yielded high results in the literature, effectively supporting manual coding performed by humans \cite{zhou-et-al}.

\subsection{Models based on neural networks}

Deep learning-based methods have significantly improved the disease coding task in recent years. \rojo{The advantage of using these models is that the healthcare-specific domain knowledge is no longer needed for the manual development of complex rules.} In contrast, these methods can automatically build features powerful enough to capture the relationships between clinical texts and their respective codes.

Most proposed systems are based on posing the problem as a multi-label text classification task \cite{karimi-etal-2017-automatic, mullenbach-etal-2018-explainable, YU2019103114, cao-etal-2020-hypercore}. Thus, the algorithm's input is text, while the output can be one or more codes associated with diseases. Unlike traditional text classification problems, this problem is considered extreme since the number of possible labels increases to thousands (depending on the terminology).

The main disadvantage of this approach is that manual coding requires incorporating context-specific rules, such as code priority, medical assumptions, code definition, and synonyms, among other types of information, to improve system performance. In the case of deep learning, this is not considered since the systems are commonly created using an end-to-end approach, meaning that no human knowledge is involved when creating the features or making the predictions.

To solve the previous problem, we followed another approach used in the literature, which consists of mixing the previous ideas using two sequential steps; the first one uses deep learning algorithms, while the second allows us to incorporate medical knowledge into the computational system. Firstly, we used a Named Entity Recognition model for automatically recognizing sequences of words in the text which are associated with diseases. Then, each disease found is associated with its most likely ICD-10 code, a task better known as Entity Linking \cite{https://doi.org/10.48550/arxiv.2010.01165, wiegreffe-etal-2019-clinical}. Nowadays, the most commonly used methods for solving the NER task are based on deep neural networks such as transformers-based models or recurrent neural networks, while a frequent technique for assigning codes is to use distance algorithms or search engines to compare the diseases found with the code descriptions of the terminology.

\subsection{Commercial Systems}

A handful of commercial products offer information extraction from clinical data, including automatic coding. These products usually are delivered as services and offered by leading cloud providers such as Amazon Web Services with Amazon Comprehend Medical\footnote{\url{https://aws.amazon.com/comprehend/medical/}}, Google Cloud with Google Cloud Healthcare Data Engine\footnote{\url{https://cloud.google.com/healthcare}} and Microsoft Azure with Azure Cognitive Service for Language\footnote{\url{https://azure.microsoft.com/en-in/products/cognitive-services/language-service}}. The problem with these services is that they do not offer automatic coding for languages other than English.

Data privacy concerns may arise from using this third-party software to extract patients' information. Some healthcare providers may prohibit sending data to systems outside the primary source due to potential cybersecurity issues.

\section{Data and Methods}
\label{data}

\rojo{The Chilean Waiting List is \jd{characteristic of the} \rojo{the public healthcare system}. This list arises due to the high demand for medical care and the limited capacity of the public health system to meet it. Entry on the waiting list begins when a patient goes to primary care or secondary care physician to treat pathology. The patient has two possible paths: if the pathology is included in the \rojo{``Garantías Explícitas en Salud'' (GES)} program, the patient enters a process where his or her health problem is assured a maximum waiting time for medical attention. If the GES program does not cover the pathology, the \jd{referral is classified in one of these five options}: New Specialty Consultations (CNE), Follow-up Consultations (CCE), Diagnostic Procedures (Proc), Surgical Intervention (IQ) and Complex Surgical Intervention (IQC). In any of these alternatives, the patient is placed on a waiting list and must wait a variable amount of time to receive medical attention from a specialist.}

The Chilean Waiting List comprises 25,374,491 referrals, divided into five categories: 18,716,629 correspond to CNE type referrals, 4,391,257 to Proc type referrals, 2,222,545 to IQ type referrals, 39,266 to CCE type referrals, and finally, 4,794 to IQC type referrals. \rojo{In particular, this work will focus on CNE-type referrals.}

\rojo{Within the Chilean Waiting database}, 73 attributes are separated into two main types of sets. The first set corresponds to the attributes associated with the person (date of birth, sex, national identifier). In contrast, the second set corresponds to the administrative information associated with the referral given to the person (date of admission, date of discharge, the benefit provided, specialty, diagnostic suspicion, and diagnostic confirmation).

For the analysis of the diagnoses present in the referrals, two free-text attributes representing medical diagnoses are considered: diagnostic suspicion and diagnostic confirmation. Table \ref{specialty-frequency} shows the frequency of referrals according to medical specialty, \rojo{while Table \ref{statistics} shows corpus statistics of the texts analyzed.} 



\begin{table}[!h]
\centering
\begin{tabular}{lll}
\hline
Specialty            & Referrals & \multicolumn{1}{l}{\begin{tabular}[c]{@{}l@{}}Relative\\ Freq. (\%)\end{tabular}} \\ \hline
Ophthalmology        & 3,352,203 & 17.91                                                                            \\
Otorhinolaryngology  & 1,270,563 & 6.79                                                                             \\
Traumatology         & 1,066,814 & 5.70                                                                             \\
Gynecology           & 991,166   & 5.30                                                                             \\
General Surgery      & 982,500   & 0.05                                                                             \\
Dermatology          & 762,758   & 4.08                                                                             \\
Internal Medicine    & 703,844   & 3.76                                                                             \\
Endodontics          & 662,607   & 3.54                                                                             \\
Removable Prosthesis & 652,604   & 3.49                                                                             \\
Urology              & 605,425   & 3.23                                                                             \\ \hline
\end{tabular}
\caption{Top 10 specialties with the highest presence on the Chilean Waiting List for a medical appointment.}
\label{specialty-frequency}
\end{table}

\begin{table*}[!h]
\centering
\begin{tabular}{llll}
\hline
Specialty                & \begin{tabular}[c]{@{}l@{}}Number of\\  tokens (std)\end{tabular} & \begin{tabular}[c]{@{}l@{}}Number of\\ sentences (std)\end{tabular} & \begin{tabular}[c]{@{}l@{}}Tokens per\\ sentence\end{tabular} \\ \hline
Infectology  & 28.59 (48.04)                                                         & 1.50 (1.61)                                                         & 18.94                                                         \\
Medical Oncology            & 20.09 (42.12)                                                   & 1.15 (0.66)                                                            & 17.40                                                         \\
Diabetology     & 19.03 (33.07)                                                        & 1.33 (1.31)                                                           & 14.22                                                         \\
Pediatric Rheumatology                & 15.05 (30.95)                                                         & 1.19 (0.75)                                                            & 12.61                                                         \\
Oral Pathology     & 14.54 (24.51)                                                        & 1.17 (0.69)                                                           & 12.34                                                         \\
Neonatology              & 12.92 (31.31)                                                        & 1.10 (0.55)                                                           & 11.64                                                         \\
Pediatric Hemato-Oncology      & 12.70 (26.74)                                                        & 1.16 (0.75)                                                           & 10.94                                                         \\
Pediatric Plastic Surgery       & 17.51 (16.33)                                                        & 1.25 (0.51)                                                           & 10.88                                                         \\
Pediatric Gynecology                & 13.01 (22.61)                                                        & 1.22 (0.82)                                                           & 10.61                                                         \\
Pediatric Traumatology & 12.27 (18.39)                                                        & 1.16 (0.74)                                                           & 10.55   \\ \hline                                             
\end{tabular}
\caption{Top 10 specialties with the highest number of tokens per sentence on average in Chilean Waiting List.}
\label{statistics}
\end{table*}

We used 10,000 referrals from the historical Chilean Waiting List to train the NER module for disease recognition. As detailed in \cite{baez-etal-2020-chilean, 10.1145/3498324}, these referrals were previously consolidated by a team of clinical experts, thus constituting the so-called Chilean Waiting list corpus. In addition, we performed rounds of evaluation of the NER performance, identifying diseases that the model could not identify. Thus, these diseases were incorporated as new examples of the model training process.



\section{Proposed System}

To code the narratives, we first used a NER model to automatically recognize sequences of words in the text associated with diseases. Then, each disease found is associated with its most likely ICD-10 code through a search engine. Figure \ref{figure:system} shows an overview of our proposed system.

\subsection{NER Model}

\begin{figure*}[!h]
  \centering
  \includegraphics[width=1.0\linewidth]{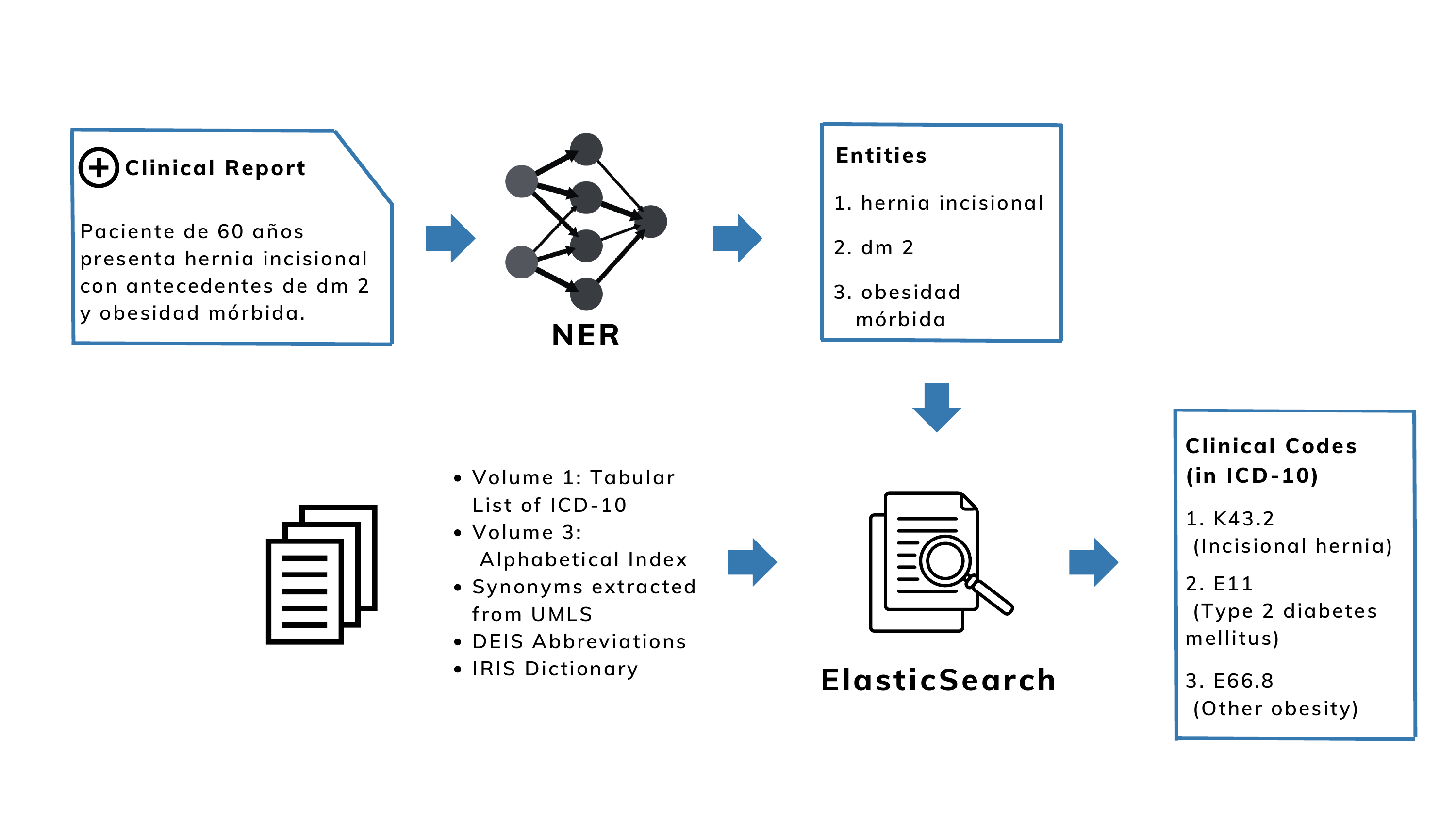}
  \caption{Overview of the proposed disease coding system. \\ \textsc{Translation} Clinical report: A 60-year-old patient presented with an incisional hernia with a history of dm 2 and morbid obesity. Entities: 1. incisional hernia, 2. dm 2, 3. morbid obesity.}
  \label{figure:system}
\end{figure*}

As shown in \rojo{Figure \ref{figure:system}}, the input of our system is the referral written by the physician in an unstructured format. These texts are used as input for the automatic disease recognition model. In particular, this NER model is based on the work proposed in \citep{rojas-etal-2022-simple}, where a simple but highly effective architecture for medical entity recognition is introduced. \rojo{This model, named Multiple LSTM-CRF (MLC), is a deep neural network system composed of three main modules, emphasizing the impact of using domain-specific contextualized embeddings.}

The first layer of the MLC approach, the ``stacked embedding layer'', transforms the texts associated with the diagnoses into a vector representation using character-level contextual embeddings and static word embeddings, both trained in the clinical domain. Then, in the encoding layer, a recurrent neural network is used to obtain long-distance dependencies between words in the sentence, thus obtaining a better context to improve the previous layer's representations. Finally, the classification layer assigns the most probable label to each word in the diagnosis using the CRF algorithm, identifying which parts of the text correspond to the beginning and end of a disease.

Regarding the experimental setup, the disease model was trained to 150 epochs using an SGD optimizer with mini-batches of size 32 and a learning rate of 0.1. As mentioned, to encode sentences, we used two types of representations; a 300-dimensional word embedding model trained on the Chilean Waiting List corpus\footnote{\url{https://zenodo.org/record/3924799}} and character-level contextualized embeddings retrieved from the Clinical Flair model \cite{rojas-etal-2022-clinical}. To implement the model and perform our experiments, we used the Flair framework, widely used by the NLP research community \cite{akbik2019flair}.

\subsection{Search Engine}

The output of the NER step is a list containing all the diseases mentioned in the referral. This second module aims to assign an ICD-10 code to each disease found, which can be used later for clinical decisions or management. The assignment of the ICD-10 code is done through a search engine tool based on Elasticsearch\footnote{Registered trademark of Elasticsearch B.V. Available at \url{https://www.elastic.co/elasticsearch/}}, an open-source search and analytics engine. This system can assign similarities between the mention of the disease and each of the codes of the ICD-10 tabular list. 

Unlike the algorithms of distance comparison between words, this search engine has an index that contains each of the ICD-10 diseases represented through a series of synonymous sentences extracted from different sources of information, simulating in a better way the process followed by clinical experts to determine the code of a disease. 

For example, in the index, the code ``K02.2'' contains the \rojo{canonical code description} ``Caries of cementum'' and multiple synonymous definitions, such as ``Cement caries'' and ``Root caries''. This is important as disease mentions found in unstructured diagnoses are rarely equivalent to the exact definition.

The sources of information used for the extraction of synonymous disease definitions were as follows:

\begin{description}
    \item[Tabular list of ICD-10 terminology:] This is the basis of the index, which tells us which codes we will assign to the disease mentions. 
    \item[Alphabetical index of ICD-10 terminology:] The guide for the manual assignment of codes to diseases and was obtained using the ``web scraping'' technique from the website of the Spanish Ministry of Health \footnote{\url{https://eciemaps.mscbs.gob.es/ecieMaps/browser/index_10_2008.html}}.
    \item[IRIS dictionary:] It maps natural language sentences to an ICD-10 code. This dictionary was built from the mortality coding rounds conducted in the Chilean Department of Statistics and Health Information.
    \item[UMLS:] Spanish definitions from multiple vocabularies were extracted from the metatresaurus database.
    \item[DEIS abbreviations:] Manually constructed list of abbreviations and their expansions. 
\end{description}

\subsection{Experiments}

In our experiments, we measure how well the predictions made by the model fit compared to the decisions made by clinical experts. In this way, a subset of the referrals described in Section \ref{data} was selected to be manually coded by a team of two clinical coders. The manual annotation process and system validation steps are provided below.

\subsubsection{Manual coding}


\rojo{The clinical experts carried out the annotation process using Excel software. For this purpose, a file containing a unique identifier for each referral, the associated diagnostic suspicion, and a blank column for the actual coding was provided to the coders. This way, the expert coders identified disease codes in 1,188 clinical narratives from the Chilean Waiting List for a new specialty.}

It is important to mention that in this process, codes were identified at the referral level, not at the entity level; therefore, it is not possible to determine the performance of the NER model in this experiment. In future work, specialized software such as \rojo{INCEpTION}, could be used, as proposed in the work of \cite{baez-etal-2020-chilean}. This software would make it possible to identify which parts of the text refer to diseases. On the other hand, only diseases were coded, but future research could extend it to new entity types, such as clinical procedures or clinical findings.

\subsubsection{Metric}


The Mean Average Precision (MAP) metric is used to evaluate the performance of our coding system. This metric is widely used in works that address the same automatic coding task. This metric is defined as follows:

\begin{equation}
    AveP = \frac{\sum (P(k)\cdot rel(k))}{\textup{number of relevant documents}},
\end{equation}
\jd{where} P(k) represents the precision at position k, and rel(k) is an indicator function equal to 1 if the element in rank k is a relevant document and 0 otherwise.

The MAP is computed using the Python implementation of the TREC evaluation tool, \texttt{trectools}, by \cite{palotti2019}, where an adaptation was applied, in which the coded diagnoses have to be ordered based on a ranking, which for this work is considered the order in which the mention was found and subsequently the code was assigned.

\section{Results}

\begin{table}[!h]
	\centering
	\begin{threeparttable}
		\begin{tabular}{ll}
			\hline
			Orthodontics (3.28) & Obstetrics (1.01)       \\
			Endodontics (2.52)  & Plastic Surgery (0.93)  \\
			Oral Rehab. (1.26)  & Pediatrics (0.84)       \\
			Nephrology (1.26)   & P. Physical Med. (0.84) \\
			P. Dentistry (1.26) & Dental Operatory (0.34) \\
			Psychiatry (1.18)   & Maxillo. Radiol. (0.34) \\
			P. Urology (1.01)   & STI (0.084)             \\
			\hline
		\end{tabular}
		\begin{tablenotes}[para,flushleft]
			\tiny P. = Pediatric, Maxillo. Radiol. = Maxillofacial Radiology, STI = Sexually transmitted infections.%
		\end{tablenotes}
	\end{threeparttable}
	\caption{Specialties with a perfect MAP score. The relative frequency (in percentage) of referrals in the dataset is in parentheses.}
	\label{10_best}
\end{table}

\begin{table*}[!h]
\centering
\begin{tabular}{lll}
\hline
Specialty                              & \begin{tabular}[c]{@{}l@{}}MAP at \\ Category Level\end{tabular} & \begin{tabular}[c]{@{}l@{}}Relative freq. \\ in \%\end{tabular} \\ \hline
Neurology                        & 0.68                                                          & 2.10                                                            \\
Immunology                             & 0.67                                                          & 0.84                                                            \\
Geriatrics                               & 0.60                                                          & 0.93                                                            \\
Pediatric Gastroenterology             & 0.58                                                          & 0.84                                                            \\
Cardiothoracic surgery                         & 0.53                                                          & 0.84                                                            \\
Radiation therapy                  & 0.50                                                          & 0.84                                                            \\
Pediatric Family Medicine             & 0.45                                                          & 0.84                                                            \\
Hematology                             & 0.42                                                          & 1.18                                                            \\
Diabetology                            & 0.35                                                          & 1.01                                                            \\
Pediatric Traumatology & 0.29                                                          & 0.93                                                            \\ \hline
\end{tabular}
\caption{Top 10 worst scores according to the specialties.}
\label{10_worst}
\end{table*}

\subsection{Coding Performance}

The ICD-10 consists of a solitary coded catalog composed of categories with three characters, each of which can be additionally subdivided into as many as ten subcategories of four characters.

\rojo{We computed the MAP metric over the test set at the category (e.g. K02) and subcategory (e.g. K02.2) levels. We achieved a MAP of 0.83 for the category and 0.63 for the subcategory level.}


To underline the difficulty of achieving outstanding results in coding, we analyzed the results obtained only by clinical experts. The expert coders achieved an agreement MAP of 0.75 for subcategory and 0.83 for category level. Several reasons, such as the subjectivity in clinical judgment, the complexity of coding guidelines, the evolving nature of medicine, the time pressure and workload, personal bias, and lack of standardization, could explain the low agreement score.


\section{Error Analysis}

To better understand the errors \rojo{made} by our coding system, we performed a granular analysis of the scores obtained among the different specialties in the corpus. Tables \ref{10_best} and \ref{10_worst} show the top 14 best and 10 worst scores according to the specialties. \rojo{We noted that in the top 14 best specialties the diagnostic suspicions registered in the referral were written straightforwardly and were specific diagnoses, such as ``lipoma'', ``caries'', and ``nephrolithiasis'', avoiding other clinical information like comorbidity, medication intake, or some other medical history.} Furthermore, it can be noted that half of these referrals are related to dental diagnosis. 

On the other hand, the top 10 worst specialties share in common that most of the diagnoses are very unspecific, with the incorporation of non-medical information such as the patient's phone number, patient's address, physician's name, the specialty the patient is referred to and information about comorbidity. Besides, several referrals are without a diagnosis but with the text ``unspecific consultation'' or ``other''.

\section{Model Deployment and Use Cases}

Due to internal regulations, we could not send patients' data to third-party systems such as cloud providers or academic supercomputing clusters \cite{ministeriosecretaríaregionaldelapresidencia1999}. For this reason, we deployed the whole coding system on-premise on a bare metal machine with a GPU compute module \rojo{(NVIDIA RTX A4000\footnote{The compute module has 16 GB of GPU memory and 6.144 CUDA cores. More information at \url{https://www.nvidia.com/en-us/design-visualization/rtx-a4000/}})} to process the coding requests from the whole department efficiently.

The complete automatic coding system was deployed as a pair of microservices running inside containers to ease portability. One container hosts the NER module and exposes an API as a web service listening to disease-mention detection requests. The other container consists of the recommended implementation of the Elasticsearch software, which also exposes its API as a web service listening to mention-coding requests.

To code the waiting list and schedule recurrent coding when new data arrives, we used the KNIME\footnote{Registered trademark of KNIME GmbH. Available at \url{https://www.knime.com/}} software, a visual-programming data mining platform. We chose this software because of its ease of use for non-expert developers. The workflow starts with the raw waiting list, which is first passed through the NER module to detect disease mentions, and then each mention is sent to the coding module to assign the most relevant code.

The automatic coding result from the workflow mentioned above is persisted on a table inside a database that stores each disease mention for each referral along with the predicted code from the system.


\section{Conclusions}
In this work, we created a nationwide system to \rojo{improve} the management of the Chilean public healthcare system. Specifically, we addressed the challenge of creating an automated system to code the diseases present in the Chilean Waiting List referrals. We developed and validated a model based on two steps: a NER model to recognize disease mentions and a search engine based on \rojo{Elasticsearch} to assign the codes to each disease. This mapping system was enriched with several terminology resources used in real life by manual coders to assign codes, thus partially simulating the pipeline followed by these professionals when solving this task.

The system \rojo{allowed us} to assign codes to 18,716,629 referrals, thus demonstrating its efficiency and effectiveness. The performance obtained in our experiments was \rojo{0.83 according to the MAP score}, which is close to the most advanced systems currently in the coding task. \rojo{The model was deployed into production in the Department of Health Statistics and Information Systems of the Ministry of Health of Chile}. 

The use of this system could be an important support for \jd{the management of waiting lists. In addition, since 75\% of the Chilean population is in the public healthcare system, the analysis of the new specialty consultations can be used for epidemiological studies, such as the one done on the incidence of psoriasis \cite{lecaros2021incidence}.}

\section*{Acknowledgements}

This work was funded by ANID Chile: Basal Funds for Center of Excellence FB210005 (CMM); Millennium Science Initiative Program ICN17\_002 (IMFD) and ICN2021\_004 (iHealth), Fondecyt grant 11201250, and National Doctoral Scholarship 21220200. We also acknowledge Daily Piedra and Marcela Carmona for their work on annotating and coding the test dataset.

\bibliography{anthology,custom}

\begin{thebibliography}{27}
\expandafter\ifx\csname natexlab\endcsname\relax\def\natexlab#1{#1}\fi

\bibitem[{Akbik et~al.(2019)Akbik, Bergmann, Blythe, Rasul, Schweter, and
  Vollgraf}]{akbik2019flair}
Alan Akbik, Tanja Bergmann, Duncan Blythe, Kashif Rasul, Stefan Schweter, and
  Roland Vollgraf. 2019.
\newblock \href {https://doi.org/10.18653/v1/N19-4010} {{FLAIR}: An easy-to-use
  framework for state-of-the-art {NLP}}.
\newblock In \emph{Proceedings of the 2019 Conference of the North {A}merican
  Chapter of the Association for Computational Linguistics (Demonstrations)},
  pages 54--59, Minneapolis, Minnesota. Association for Computational
  Linguistics.

\bibitem[{B{\'a}ez et~al.(2022)B{\'a}ez, Arancibia, Chaparro, Bucarey,
  N{\'u}{\~n}ez, and Dunstan}]{baez2022procesamiento}
Pablo B{\'a}ez, Antonia~Paz Arancibia, Mat{\'\i}as~Ignacio Chaparro, Tom{\'a}s
  Bucarey, Fredy N{\'u}{\~n}ez, and Jocelyn Dunstan. 2022.
\newblock Procesamiento de lenguaje natural para texto cl{\'\i}nico en
  espa{\~n}ol: el caso de las listas de espera en chile.
\newblock \emph{Revista M{\'e}dica Cl{\'\i}nica Las Condes}, 33(6):576--582.

\bibitem[{B\'{a}ez et~al.(2022)B\'{a}ez, Bravo-Marquez, Dunstan, Rojas, and
  Villena}]{10.1145/3498324}
Pablo B\'{a}ez, Felipe Bravo-Marquez, Jocelyn Dunstan, Mat\'{\i}as Rojas, and
  Fabi\'{a}n Villena. 2022.
\newblock \href {https://doi.org/10.1145/3498324} {Automatic extraction of
  nested entities in clinical referrals in spanish}.
\newblock \emph{ACM Trans. Comput. Healthcare}, 3(3).

\bibitem[{B{\'a}ez et~al.(2020)B{\'a}ez, Villena, Rojas, Dur{\'a}n, and
  Dunstan}]{baez-etal-2020-chilean}
Pablo B{\'a}ez, Fabi{\'a}n Villena, Mat{\'\i}as Rojas, Manuel Dur{\'a}n, and
  Jocelyn Dunstan. 2020.
\newblock \href {https://doi.org/10.18653/v1/2020.clinicalnlp-1.32} {The
  {C}hilean waiting list corpus: a new resource for clinical named entity
  recognition in {S}panish}.
\newblock In \emph{Proceedings of the 3rd Clinical Natural Language Processing
  Workshop}, pages 291--300, Online. Association for Computational Linguistics.

\bibitem[{Barros et~al.(2022)Barros, Rojas, Dunstan, and
  Abeliuk}]{barros2022divide}
Jose Barros, Mat{\'\i}as Rojas, Jocelyn Dunstan, and Andres Abeliuk. 2022.
\newblock Divide and conquer: An extreme multi-label classification approach
  for coding diseases and procedures in spanish.
\newblock In \emph{Proceedings of the 13th International Workshop on Health
  Text Mining and Information Analysis (LOUHI)}, pages 138--147.

\bibitem[{Cao et~al.(2020)Cao, Chen, Liu, Zhao, Liu, and
  Chong}]{cao-etal-2020-hypercore}
Pengfei Cao, Yubo Chen, Kang Liu, Jun Zhao, Shengping Liu, and Weifeng Chong.
  2020.
\newblock \href {https://doi.org/10.18653/v1/2020.acl-main.282} {{H}yper{C}ore:
  Hyperbolic and co-graph representation for automatic {ICD} coding}.
\newblock In \emph{Proceedings of the 58th Annual Meeting of the Association
  for Computational Linguistics}, pages 3105--3114, Online. Association for
  Computational Linguistics.

\bibitem[{Dalianis(2018)}]{dalianis2018}
Hercules Dalianis. 2018.
\newblock \href {https://doi.org/10.1007/978-3-319-78503-5} {\emph{Clinical
  Text Mining}}, first edition.
\newblock Springer International Publishing.

\bibitem[{Dong et~al.(2022)Dong, Falis, Whiteley, Alex, Matterson, Ji, Chen,
  and Wu}]{Dong2022}
Hang Dong, Mat{\'{u}}{\v{s}} Falis, William Whiteley, Beatrice Alex, Joshua
  Matterson, Shaoxiong Ji, Jiaoyan Chen, and Honghan Wu. 2022.
\newblock \href {https://doi.org/10.1038/s41746-022-00705-7} {Automated
  clinical coding: what, why, and where we are?}
\newblock \emph{npj Digital Medicine}, 5(1).

\bibitem[{Farkas and Szarvas(2008)}]{Farkas2008AutomaticCO}
Rich{\'{a}}rd Farkas and Gy\"{o}rgy Szarvas. 2008.
\newblock \href {https://doi.org/10.1186/1471-2105-9-s3-s10} {Automatic
  construction of rule-based {ICD}-9-{CM} coding systems}.
\newblock \emph{{BMC} Bioinformatics}, 9(S3).

\bibitem[{Karimi et~al.(2017)Karimi, Dai, Hassanzadeh, and
  Nguyen}]{karimi-etal-2017-automatic}
Sarvnaz Karimi, Xiang Dai, Hamed Hassanzadeh, and Anthony Nguyen. 2017.
\newblock \href {https://doi.org/10.18653/v1/W17-2342} {Automatic diagnosis
  coding of radiology reports: A comparison of deep learning and conventional
  classification methods}.
\newblock In \emph{{B}io{NLP} 2017}, pages 328--332, Vancouver, Canada,.
  Association for Computational Linguistics.

\bibitem[{Kong(2019)}]{kong}
Hyoun-Joong Kong. 2019.
\newblock \href {https://doi.org/10.4258/hir.2019.25.1.1} {Managing
  unstructured big data in healthcare system}.
\newblock \emph{Healthcare Informatics Research}, 25:1.

\bibitem[{Kraljevic et~al.(2021)Kraljevic, Searle, Shek, Roguski, Noor, Bean,
  Mascio, Zhu, Folarin, Roberts, Bendayan, Richardson, Stewart, Shah, Wong,
  Ibrahim, Teo, and Dobson}]{https://doi.org/10.48550/arxiv.2010.01165}
Zeljko Kraljevic, Thomas Searle, Anthony Shek, Lukasz Roguski, Kawsar Noor,
  Daniel Bean, Aurelie Mascio, Leilei Zhu, Amos~A. Folarin, Angus Roberts,
  Rebecca Bendayan, Mark~P. Richardson, Robert Stewart, Anoop~D. Shah,
  Wai~Keong Wong, Zina Ibrahim, James~T. Teo, and Richard~J.B. Dobson. 2021.
\newblock \href {https://doi.org/https://doi.org/10.1016/j.artmed.2021.102083}
  {Multi-domain clinical natural language processing with medcat: The medical
  concept annotation toolkit}.
\newblock \emph{Artificial Intelligence in Medicine}, 117:102083.

\bibitem[{Lecaros et~al.(2021)Lecaros, Dunstan, Villena, Ashcroft, Parisi,
  Griffiths, H{\"a}rtel, Maul, and De~la Cruz}]{lecaros2021incidence}
C~Lecaros, J~Dunstan, F~Villena, DM~Ashcroft, R~Parisi, CEM Griffiths,
  S~H{\"a}rtel, JT~Maul, and C~De~la Cruz. 2021.
\newblock The incidence of psoriasis in chile: an analysis of the national
  waiting list repository.
\newblock \emph{Clinical and Experimental Dermatology}, 46(7):1262--1269.

\bibitem[{{Ministerio de Salud de Chile}(2011)}]{normaTecnica}
{Ministerio de Salud de Chile}. 2011.
\newblock \href {www.supersalud.gob.cl/664/w3-propertyvalue-6249.html} {Norma
  técnica parael registro de las listas de espera}.

\bibitem[{{Ministerio Secretaría Regional de la
  Presidencia}(1999)}]{ministeriosecretaríaregionaldelapresidencia1999}
{Ministerio Secretaría Regional de la Presidencia}. 1999.
\newblock \href
  {https://www.bcn.cl/leychile/navegar?idNorma=141599&idVersion=2020-08-26}
  {Ley 19628 sobre protección de la vida privada}.

\bibitem[{Mullenbach et~al.(2018)Mullenbach, Wiegreffe, Duke, Sun, and
  Eisenstein}]{mullenbach-etal-2018-explainable}
James Mullenbach, Sarah Wiegreffe, Jon Duke, Jimeng Sun, and Jacob Eisenstein.
  2018.
\newblock \href {https://doi.org/10.18653/v1/N18-1100} {Explainable prediction
  of medical codes from clinical text}.
\newblock In \emph{Proceedings of the 2018 Conference of the North {A}merican
  Chapter of the Association for Computational Linguistics: Human Language
  Technologies, Volume 1 (Long Papers)}, pages 1101--1111, New Orleans,
  Louisiana. Association for Computational Linguistics.

\bibitem[{Palotti et~al.(2019)Palotti, Scells, and Zuccon}]{palotti2019}
Jo\~{a}o Palotti, Harrisen Scells, and Guido Zuccon. 2019.
\newblock \href {https://doi.org/10.1145/3331184.3331399} {Trectools: An
  open-source python library for information retrieval practitioners involved
  in trec-like campaigns}.
\newblock In \emph{Proceedings of the 42nd International ACM SIGIR Conference
  on Research and Development in Information Retrieval}, SIGIR'19, page
  1325–1328, New York, NY, USA. Association for Computing Machinery.

\bibitem[{Rojas et~al.(2022{\natexlab{a}})Rojas, Bravo-Marquez, and
  Dunstan}]{rojas-etal-2022-simple}
Matias Rojas, Felipe Bravo-Marquez, and Jocelyn Dunstan. 2022{\natexlab{a}}.
\newblock \href {https://aclanthology.org/2022.coling-1.184} {Simple yet
  powerful: An overlooked architecture for nested named entity recognition}.
\newblock In \emph{Proceedings of the 29th International Conference on
  Computational Linguistics}, pages 2108--2117, Gyeongju, Republic of Korea.
  International Committee on Computational Linguistics.

\bibitem[{Rojas et~al.(2022{\natexlab{b}})Rojas, Dunstan, and
  Villena}]{rojas-etal-2022-clinical}
Mat{\'\i}as Rojas, Jocelyn Dunstan, and Fabi{\'a}n Villena. 2022{\natexlab{b}}.
\newblock \href {https://doi.org/10.18653/v1/2022.clinicalnlp-1.9} {Clinical
  flair: A pre-trained language model for {S}panish clinical natural language
  processing}.
\newblock In \emph{Proceedings of the 4th Clinical Natural Language Processing
  Workshop}, pages 87--92, Seattle, WA. Association for Computational
  Linguistics.

\bibitem[{Stanfill et~al.(2010)Stanfill, Williams, Fenton, Jenders, and
  Hersh}]{coding-review}
Mary Stanfill, Margaret Williams, Susan Fenton, Robert Jenders, and William
  Hersh. 2010.
\newblock \href {https://doi.org/10.1136/jamia.2009.001024} {A systematic
  literature review of automated clinical coding and classification systems}.
\newblock \emph{Journal of the American Medical Informatics Association :
  JAMIA}, 17:646--51.

\bibitem[{Teng et~al.(2022)Teng, Liu, Li, Zhang, Li, and Zhao}]{teng-et-al}
Fei Teng, Yiming Liu, Tianrui Li, Yi~Zhang, Shuangqing Li, and Yue Zhao. 2022.
\newblock \href {https://doi.org/10.1109/TKDE.2022.3148267} {A review on deep
  neural networks for icd coding}.
\newblock \emph{IEEE Transactions on Knowledge and Data Engineering}, pages
  1--1.

\bibitem[{Wiegreffe et~al.(2019)Wiegreffe, Choi, Yan, Sun, and
  Eisenstein}]{wiegreffe-etal-2019-clinical}
Sarah Wiegreffe, Edward Choi, Sherry Yan, Jimeng Sun, and Jacob Eisenstein.
  2019.
\newblock \href {https://doi.org/10.18653/v1/W19-5028} {Clinical concept
  extraction for document-level coding}.
\newblock In \emph{Proceedings of the 18th BioNLP Workshop and Shared Task},
  pages 261--272, Florence, Italy. Association for Computational Linguistics.

\bibitem[{{World Health Organization}(2015)}]{World_Health_Organization2015-tx}
{World Health Organization}. 2015.
\newblock \href {https://apps.who.int/iris/handle/10665/246208}
  {\emph{International statistical classification of diseases and related
  health problems}}, 10th revision, fifth edition, 2016 edition.
\newblock {World Health Organization}.

\bibitem[{{World Health Organization}(2023)}]{cie11}
{World Health Organization}. 2023.
\newblock \href {https://icdcdn.who.int/icd11referenceguide/en/html/index.html}
  {\emph{International statistical classification of diseases and related
  health problems}}, 11th revision edition.
\newblock {World Health Organization}.

\bibitem[{Yan et~al.(2022)Yan, Fu, Liu, Zhang, Gao, Wu, and Li}]{YAN2022161}
Chenwei Yan, Xiangling Fu, Xien Liu, Yuanqiu Zhang, Yue Gao, Ji~Wu, and Qiang
  Li. 2022.
\newblock \href {https://doi.org/https://doi.org/10.1016/j.imed.2022.03.003} {A
  survey of automated international classification of diseases coding:
  development, challenges, and applications}.
\newblock \emph{Intelligent Medicine}, 2(3):161--173.

\bibitem[{Yu et~al.(2019)Yu, Li, Liu, Fei, Wu, and Wang}]{YU2019103114}
Ying Yu, Min Li, Liangliang Liu, Zhihui Fei, Fang-Xiang Wu, and Jianxin Wang.
  2019.
\newblock \href {https://doi.org/https://doi.org/10.1016/j.jbi.2019.103114}
  {Automatic icd code assignment of chinese clinical notes based on multilayer
  attention birnn}.
\newblock \emph{Journal of Biomedical Informatics}, 91:103114.

\bibitem[{Zhou et~al.(2020)Zhou, Cheng, Ou, and Huang}]{zhou-et-al}
Lingling Zhou, Cheng Cheng, Dong Ou, and Hao Huang. 2020.
\newblock \href {https://doi.org/10.1186/s12911-020-1085-4} {Construction of a
  semi-automatic icd-10 coding system}.
\newblock \emph{BMC Medical Informatics and Decision Making}, 20.

\end{thebibliography}
\bibliographystyle{acl_natbib}

\end{document}